\documentclass[10pt,letterpaper]{article}
\PassOptionsToPackage{numbers,sort&compress}{natbib}
\usepackage[T1]{fontenc}
\usepackage{arxiv}
\renewcommand{\shorttitle}{CAST: Paired-Depth Generative Design and Shape Optimization}
\renewcommand{\undertitle}{}
\renewcommand{\headeright}{}
\usepackage{etoolbox}
\makeatletter
\patchcmd{\@maketitle}{\textsc{\undertitle}\\}{}{}{\PackageError{CAST}{Title label removal failed}{}}
\makeatother
\usepackage{mathtools,amssymb,bm,graphicx,xcolor,array,dcolumn,booktabs,tabularx}
\usepackage{microtype,algorithm,algpseudocode,placeins,natbib}
\usepackage[hidelinks]{hyperref}
\usepackage{doi}
\usepackage[font=small,labelfont=bf]{caption}
\newcolumntype{Y}{>{\raggedright\arraybackslash}X}
\hypersetup{
  pdfauthor={Jihoon Kim, Yongmin Kwon, and Namwoo Kang},
  pdftitle={CAST: Cast-Admissible Generative Design and Shape Optimization with Paired Depth Images},
  pdfkeywords={deep generative design, design for manufacturing, die casting, shape representation},
  pdfsubject={Paired depth fields for undercut-free deep generative design and shape optimization}
}
\title{CAST: Cast-Admissible Generative Design and Shape Optimization with Paired Depth Images}
\author{\normalfont Jihoon Kim$^{1,3,\dagger}$\quad Yongmin Kwon$^{2,3,\dagger}$\quad Namwoo Kang$^{1,3,*}$ \\
{\small\normalfont $^{1}$ Department of Mechanical Engineering,} \\
{\small\normalfont Korea Advanced Institute of Science and Technology, Daejeon 34141, Republic of Korea} \\
{\small\normalfont $^{2}$ Cho Chun Shik Graduate School of Mobility,} \\
{\small\normalfont Korea Advanced Institute of Science and Technology, Daejeon 34051, Republic of Korea} \\
{\small\normalfont $^{3}$ Narnia Labs, Daejeon 34051, Republic of Korea} \\
{\small\normalfont jihoon.kim@kaist.ac.kr\quad kymin1002@kaist.ac.kr\quad nwkang@kaist.ac.kr} \\
{\small\normalfont $^{\dagger}$ Jihoon Kim and Yongmin Kwon contributed equally to this work.} \\
{\small\normalfont $^{*}$ Corresponding author: Namwoo Kang}}
\date{}

\begin{document}
\maketitle
\begin{abstract}
Deep generative design can produce complex geometries, but unconstrained representations may create undercuts that prevent release from a two-part mold.
We present CAST, which reformulates three-dimensional shape generation under a selected mold-opening constraint as a two-dimensional image generation problem using paired top and bottom depth images.
Each valid depth pair defines one material interval along a selected mold-opening direction, so reconstructions exclude undercuts in that direction by construction.
Diffusion models generate candidate shapes in this representation, and a finite element analysis (FEA) guided optimizer refines the same two boundaries.
We evaluate diffusion-based generation with limited training data comprising 211 SimJEB bracket designs.
On SimJEB brackets, FEA-guided refinement with a volume penalty reduced compliance in 193 of 200 generated candidates. The median reduction across all candidates was 36.8 percent in the optimization FEA model.
Reconstruction of existing brackets revealed a mass--stiffness trade-off: filling occluded regions increased material volume and reduced displacement under the evaluated loads.
Paired depth images thus connect generative exploration and structural refinement in a shared design space that preserves the selected pull-direction restriction.
\end{abstract}

\section{Introduction}

Deep generative models are increasingly used in engineering design to synthesize large and diverse sets of candidate geometries from limited prior designs \citep{regenwetter2022GDreview,oh2019deep,chen2020padgan}.
Recent applications span native three-dimensional shape synthesis, topology-optimized structural components, and architected material systems \citep{shu2020three,li2022tevae,herron2024latent,chen2025multilattice}.
For these models to contribute to mass production, however, generated candidates must satisfy not only functional objectives but also the geometric requirements of the intended manufacturing process.
Although additive manufacturing can accommodate many complex proposals, its production rate and unit cost do not always suit large production volumes \citep{leary2019DesignforAM,gibson2021DesignforAM}.
At high volumes, die casting and injection molding remain economical because tooling and setup costs are spread across many parts \citep{dewhurst1989diecasting,karania2007lowvolume,favi2017hpdcost}.
Components produced by these processes must permit mold separation along a feasible draw direction, avoid undercuts, and provide sufficient draft for ejection \citep{ravi1990parting,weinst1996draw,alderighi2022moldreview}.
Functional and assembly interfaces impose additional, part-specific geometric constraints.
When these requirements are introduced only after generation, geometric repair can alter both the candidate shape and its mechanical behavior.

\begin{figure}[t]
    \centering
    \includegraphics[width=.98\linewidth,height=.68\textheight,keepaspectratio]{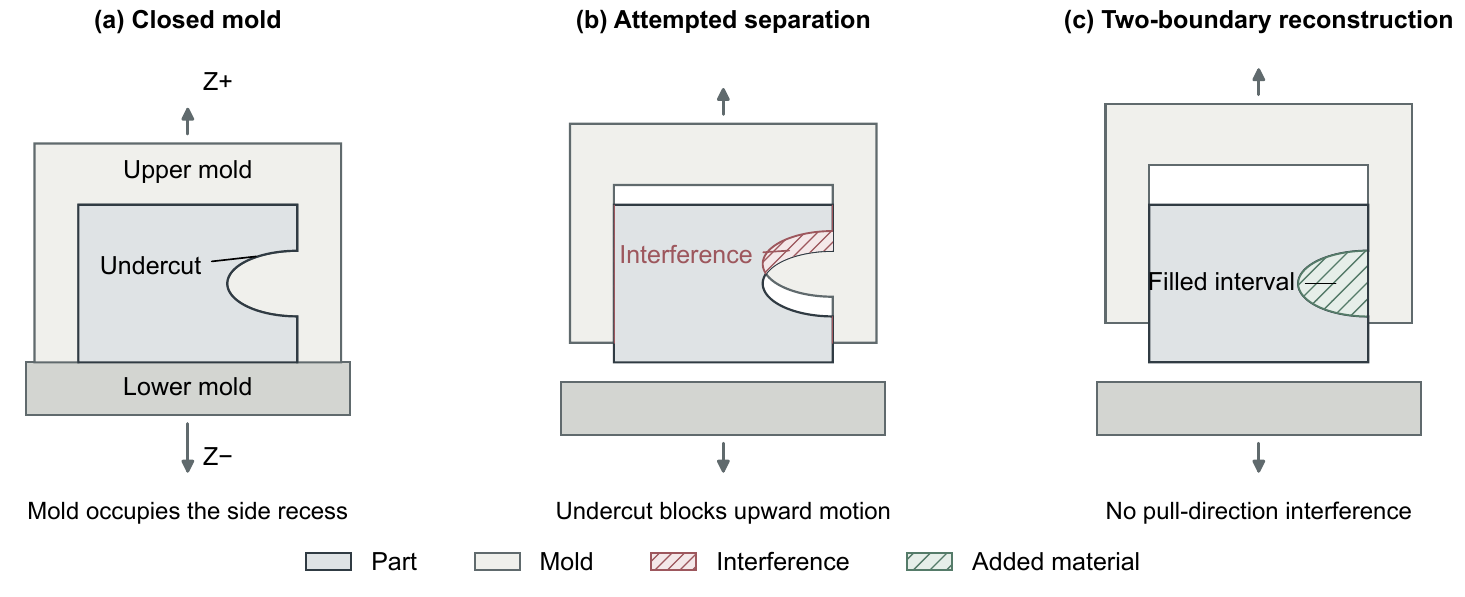}
    \caption{Undercuts and mold separation along the selected pull direction.
    (a) Mold material occupies a side recess in the part.
    (b) Attempted upward motion of the upper mold produces interference with the part, highlighted in red.
    (c) Filling the interval between the opposing boundaries removes the recess and the pull-direction interference. Sage-green hatching marks the added material.}
    \label{fig:undercut-separation}
\end{figure}

The central challenge is therefore not simply to generate complex geometry, but to make relevant manufacturing constraints native to the design space.
Geometric representations chosen for expressive flexibility do not necessarily preserve such constraints as designs move between generation, reconstruction, analysis, and refinement.
Constraint checks and learned design repair provide ways to handle these violations \citep{regenwetter2022GDreview,wang2024mcgan}.
A process-aware representation offers a more direct alternative: if every valid state already satisfies a critical geometric condition, generated candidates inherit that condition before screening or modification.
For two-part molds, one such condition is the absence of undercuts along the selected pull direction.
An undercut traps mold material and obstructs withdrawal along that direction (Fig.~\ref{fig:undercut-separation}(a,b)).

\begin{figure}[tb]
    \centering
    \includegraphics[width=.98\linewidth,height=.68\textheight,keepaspectratio]{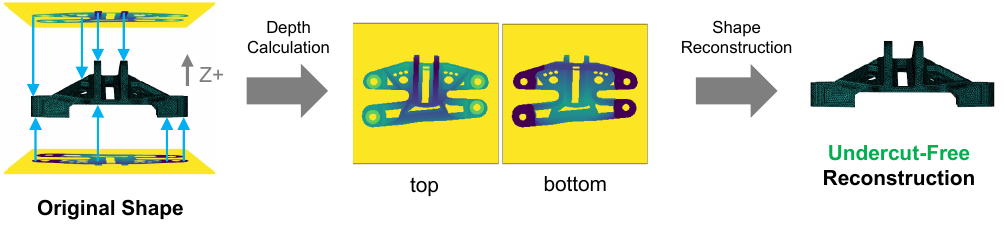}
    \caption{Depth calculation and reconstruction for paired top and bottom fields.
    Orthographic rays record first-surface distances from opposite camera planes, and the interval between the paired surfaces is filled to reconstruct a solid with at most one material interval along the selected pull direction.}
    \label{fig:depth-calc-and-shape-recon}
\end{figure}

To embed this condition in the shape representation, we use two depth images viewed from opposite sides along the selected mold-opening direction.
These images record the upper and lower surfaces, and the space between them is filled with material (Fig.~\ref{fig:depth-calc-and-shape-recon}).
This construction leaves no hidden gaps beneath overhangs that would obstruct mold withdrawal along that direction (Fig.~\ref{fig:undercut-separation}(c)).
We call shapes satisfying this geometric condition \emph{cast-admissible}.
The representation therefore excludes undercuts by construction.

\begin{figure}[t]
    \centering
    \includegraphics[width=.70\linewidth,height=.68\textheight,keepaspectratio]{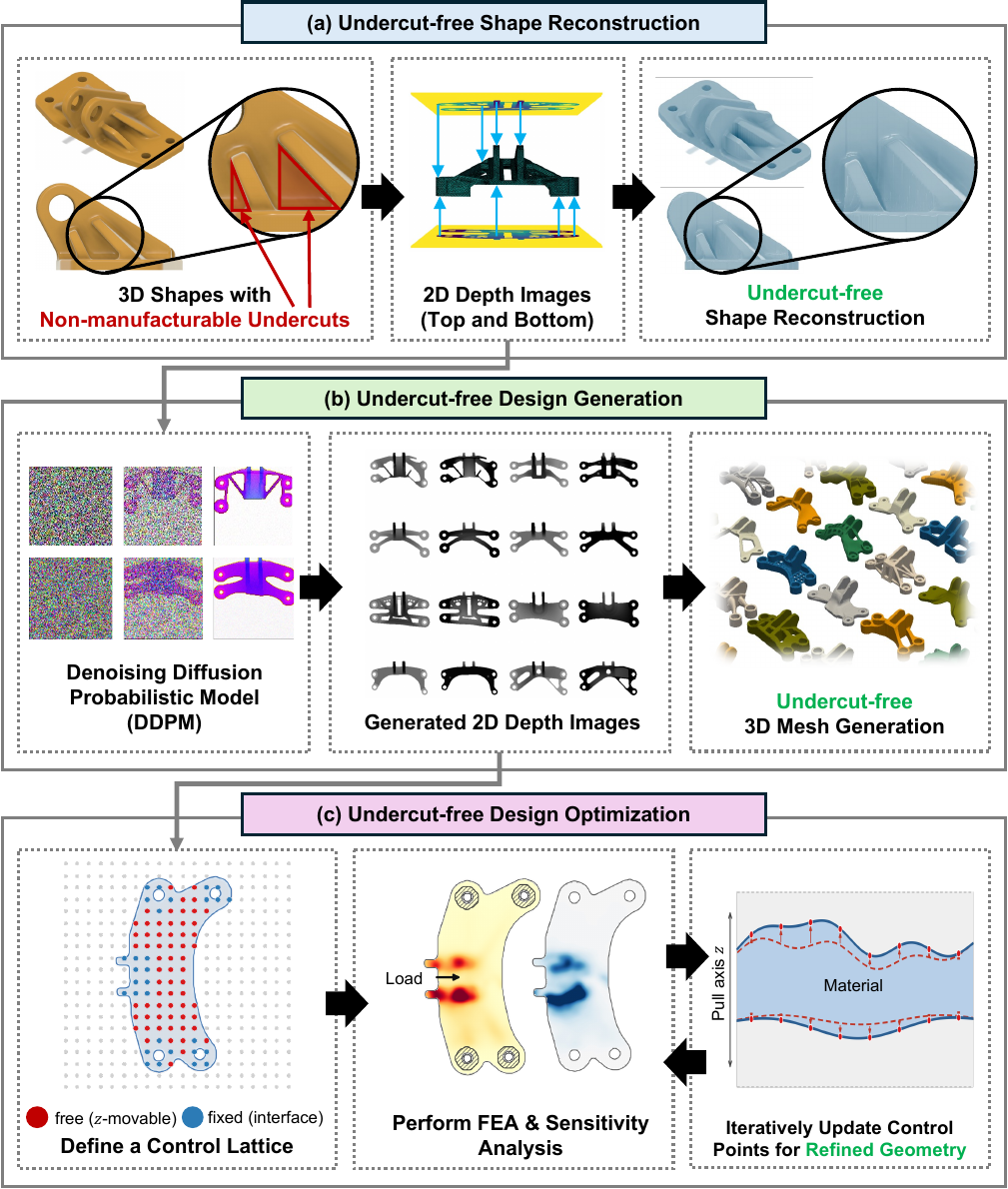}
    \caption{The CAST framework.
    (a) Valid paired depth fields reconstruct a solid without undercuts along the selected pull direction.
    (b) A diffusion model generates new paired depth fields.
    (c) FEA-guided optimization refines the same upper and lower boundaries along that direction while keeping the in-plane footprint fixed.}
    \label{fig:framework}
\end{figure}

This two-image representation enables two-dimensional generative models to propose candidate shapes without learning a dense volumetric field \citep{soltani2017multiview}.
Appropriate inductive biases can improve learning from limited data by incorporating relevant structure \citep{battaglia2018relational}.
By combining a compact representation with an explicit pull-direction constraint, CAST may therefore reduce training data requirements relative to unconstrained volumetric generation.
The same upper and lower boundaries also serve as design variables for finite element analysis (FEA) guided shape optimization, allowing generated candidates to be refined while retaining the selected pull-direction restriction.
We parameterize these boundary updates using cubic B-spline free-form deformation (FFD).
Together, the paired-depth representation, generation, and shape optimization form the CAST framework shown in Fig.~\ref{fig:framework}.

We evaluate CAST on the SimJEB bracket family \citep{whalen2021simjeb}.
Generation experiments demonstrate the creation of new candidates within the paired-depth design space, and FEA-guided optimization demonstrates structural improvement through updates to the same boundaries under a volume penalty.
Reconstruction experiments complement these results by quantifying the geometric and mass--stiffness trade-offs involved in mapping existing designs into this space.

The contributions of this work are:
\begin{itemize}
\item a paired-depth representation that reformulates pull-direction-constrained 3D shape generation as a two-dimensional image generation problem, with valid reconstructions excluding undercuts by construction;
\item a generation and FEA-guided shape optimization framework that refines the same depth boundaries used to generate candidates while retaining the selected pull-direction restriction; and
\item a structural assessment of paired-depth reconstruction that establishes its mass--stiffness trade-off on SimJEB brackets, showing how filling occluded regions increases material volume and reduces displacement across the evaluated load cases.
\end{itemize}

Section~\ref{sec:related_work} reviews related work.
Section~\ref{sec:methods} presents the paired-depth representation, generative models, and FEA-guided refinement.
Section~\ref{sec:results} evaluates reconstruction, generation, and shape optimization, followed by a discussion of the findings and limitations in Section~\ref{sec:discussion} and conclusions in Section~\ref{sec:conclusion}.

\section{Related Work}
\label{sec:related_work}

\subsection{Generative Design}
Generative design uses optimization and learned models to explore candidate geometries that satisfy performance objectives \citep{krish2011practicalGD,renner2003GA,regenwetter2022GDreview}.
Common deep generative models include variational autoencoders (VAEs), generative adversarial networks (GANs), and diffusion models \citep{kingma2013auto,goodfellow2014generative,ho2020ddpm}.
Deep learning approaches have generated topology-optimized structures \citep{guo2018indirect,oh2019deep,nie2021topologyGAN,maze2023topodiff,giannone2023diff}, lattice structures \citep{eren2024lattice}, and nanophotonic designs \citep{kim2022nanophotonics}.
Conditional GANs have also generated shapes within specified attribute ranges \citep{nobari2022rangegan}, predicted topologies using transfer learning \citep{behzadi2022gantl}, and incorporated layout and mechanical constraints through multiple discriminators \citep{almasri2024geometric}.
Deep generative and optimization methods have also produced three-dimensional shapes from limited data \citep{kwon2025three}, but such outputs target performance or diversity and are not restricted to geometry that a two-part mold can release.
Manufacturing constraints are often introduced through training data, conditioning, or post-processing \citep{rade2023TOmanufacturable,chen2019synthesizing}.
The present work instead makes one mold-release restriction explicit in the representation on which both generation and shape optimization operate.

\subsection{Design for Manufacturing}
Design for Manufacturing introduces process constraints while design freedom is still available, reducing later redesign and production cost \citep{boothroyd2010product,kuo2001dfm,formentini2022dfmreview}.
Patterson et al.~\citep{patterson2021constraints} surveyed methods for generating and enforcing process-driven manufacturability constraints across design domains.
For injection molding and die casting, relevant geometric checks include mold-opening direction, undercuts, draft, wall thickness, and interfaces \citep{boothroyd1994product,alderighi2022moldreview}.
These checks are coupled but not interchangeable: a shape can be free of undercuts in one pull direction and still violate draft or wall thickness guidance.

Casting constraints have also been incorporated into structural optimization.
Explicit parameterizations have represented solid--void boundaries along a casting direction \citep{gersborg2011casting} and optimized stiffener layout and height through directional growth \citep{liu2015hdgtp}.
Other approaches have imposed manufacturing constraints through density projections \citep{vatanabe2016manufacturing}, shadow density filters \citep{yoon2021shadow}, and density-gradient measures of undercuts \citep{qian2017undercut}.
Parting lines have been incorporated into multicomponent topology optimization \citep{zhou2019mtod}, and recent work jointly optimized cast geometry, draw directions, nonplanar parting surfaces, and side dies \citep{fu2026casting}.

Generative design research has more often treated additive manufacturing limits, where the feasible geometry differs from that of two-part molds \citep{zegard2016bridging,langelaar2016topology,hertlein2021GANAM}.
Neural networks have also been used to jointly optimize build direction, part segmentation, and topology for additive manufacturing \citep{chen2023TOforAM}.
Manufacturing-aware generation has used synthetic machinable shapes to train GANs for latent-space optimization \citep{greminger2020manufacturing} and domain rules to generate sheet metal parts \citep{wittig2026sheetgen}.
Recent work has incorporated injection molding rules into learned repair of existing designs \citep{wang2024mcgan}, and a survey of diffusion models in design and manufacturing identifies demoldability as an open constraint \citep{leng2025diffusionsurvey}.
CAST incorporates the selected pull-direction restriction into the representation used to generate and refine shapes.

\subsection{Depth Images}
Depth images store the distance from a viewing plane to the first observed surface and are widely used for three-dimensional perception and reconstruction \citep{szeliski2022computer,scharstein2002taxonomy,geng2011structured,han2013kinect}.
Layered depth images store multiple surfaces along viewing rays \citep{shade1998layered}, while layered depth-normal images support solid modeling operations \citep{wang2010ldni}.
Height fields have also been used to approximate free-form geometry for manufacturing \citep{herholz2015height}, and multi-view depth maps have supported three-dimensional shape generation with VAEs \citep{soltani2017multiview} and diffusion models \citep{wang2024mvdd}.
In CAST, two opposing orthographic depth images define the upper and lower boundaries of a solid, and filling the space between them produces a single material interval along each line parallel to the selected pull direction.

\subsection{Physics-Based Shape Optimization}
Topology optimization and learned generative models have been combined by training on optimized designs, predicting near-optimal layouts, or guiding search with learned distributions \citep{oh2019deep,yu2019deepTO,hertlein2021GANAM}.
Wang et al.~\citep{wang2023cganto} embedded compliance and volume fraction terms into conditional GAN training to generate functional designs.
Shape optimization can instead begin from a generated candidate and optimize a lower-dimensional boundary parameterization.
Cubic B-spline FFD supplies smooth boundary control through a surrounding lattice \citep{sederberg1986ffd}, and has been used to optimize the engineering performance of three-dimensional generated and augmented geometries \citep{kwon2026three}.
In CAST, FFD updates the upper and lower depth boundaries directly, enabling structural refinement within the same design representation used for generation.

CAST thus connects shape generation and FEA-guided refinement within a shared paired-depth design space that preserves the selected pull-direction restriction.

\section{Methods}\label{sec:methods}

CAST uses paired depth images to represent, generate, and refine three-dimensional shapes (Fig.~\ref{fig:framework}).
We first describe how the two depth boundaries define a solid and how they are obtained from existing shapes.
Generative models then learn to produce new depth pairs, and FEA-guided optimization refines selected candidates by moving the same boundaries.

\subsection{Two-Depth Representation}
Let the selected pull direction be the $z$ axis, and let $P_{ij}$ denote the in-plane cell associated with pixel $(i,j)$ on an $H\times W$ orthographic grid.
The two depth channels derive, rather than store separately, a foreground mask $m_{ij}$ and boundaries $z^+_{ij}$ and $z^-_{ij}$.
The represented body is
\begin{equation}\label{eq:solid}
\Omega_D=\bigcup_{i,j:m_{ij}=1}\left\{(x,y,z):(x,y)\in P_{ij},\ z^-_{ij}\le z\le z^+_{ij}\right\},
\end{equation}
where $m_{ij}=0$ omits the corresponding column.
A paired foreground column has valid depth ordering when $D_{\mathrm{top}}+D_{\mathrm{bottom}}\leq0.8$, equivalently $z^-_{ij}\leq z^+_{ij}$.
Every line parallel to $z$ then intersects $\Omega_D$ in either an empty set or one connected interval.
The no-undercut property used in this paper refers to this reconstructed body and this selected pull direction.

\subsection{Depth Calculation}\label{ssec:depth-calc}
Opposing orthographic cameras record the distance to the first surface encountered along each inward ray (Fig.~\ref{fig:depth-calc-and-shape-recon}).
Parallel projection preserves a direct mapping between each pixel and a column along the selected pull direction.
The paired distances determine the upper and lower surface coordinates, while rays that miss the shape identify empty columns.

For the SimJEB engine brackets used in this study \citep{whalen2021simjeb}, we selected the $z$ axis as the pull direction because the registered family presents its principal bracket face to that axis.
Input computer-aided design (CAD) coordinates in millimeters were mapped to a fixed normalized frame computed once over the retained 264 designs. The exact affine center and scale are provided [See Supplemental Material].
Two $256\times256$ grids use $(\widetilde{x}_i,\widetilde{y}_j)=(2i/256-1,2j/256-1)$ and camera planes $\widetilde{z}=\pm0.4$.
The first-surface distances are stored as $D_{\mathrm{top}}$ and $D_{\mathrm{bottom}}$, with 0.8 marking a missed ray.
After range mapping, $m_{ij}=1$ only when both channel values are below 0.78. Otherwise, the column is empty.
For foreground pixels, $z^+=0.4-D_{\mathrm{top}}$ and $z^-=-0.4+D_{\mathrm{bottom}}$.
Representative paired fields are shown in Fig.~\ref{fig:depth-images-demo}. The full pseudocode is provided [See Supplemental Material].

\begin{figure}[tb]
    \centering
    \includegraphics[width=95mm,height=.68\textheight,keepaspectratio]{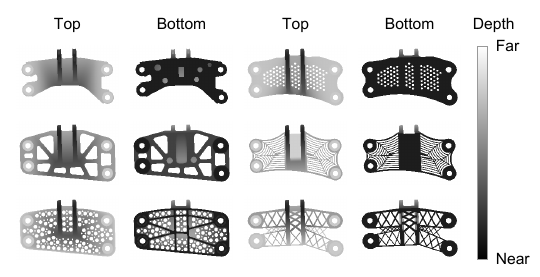}
    \caption{Paired top and bottom depth fields for six SimJEB brackets. Depth is measured in the normalized coordinate system.}
    \label{fig:depth-images-demo}
\end{figure}

\subsection{Shape Reconstruction}
At each in-plane pixel, the reconstruction algorithm forms a sub-voxel scalar field over 128 samples along $z$, fills the single interval between the top and bottom surfaces, and extracts its zero level set with marching cubes (Fig.~\ref{fig:depth-calc-and-shape-recon}).
The algorithm maps depths below 0 to 0 and depths at least 0.78 to the empty-ray value 0.8 before reconstruction.
Under valid depth ordering, the single-interval column construction excludes undercuts along the selected pull direction by construction.
It does not preserve internal cavities or features that are occluded along that direction.
The implementation-equivalent pseudocode is provided [See Supplemental Material].

\subsection{Manufacturing-Oriented Post-processing}\label{sec:postprocess}

\subsubsection{Positive Draft Angle Enforcement}
A chosen positive draft angle can also be formulated as a constraint on depth field slopes.
An illustrative projection is provided [See Supplemental Material]. It was not used in the reported experiments.

\subsubsection{Interface Feature Restoration}

\begin{figure}[tbp]
    \centering
    \includegraphics[width=80mm,height=.68\textheight,keepaspectratio]{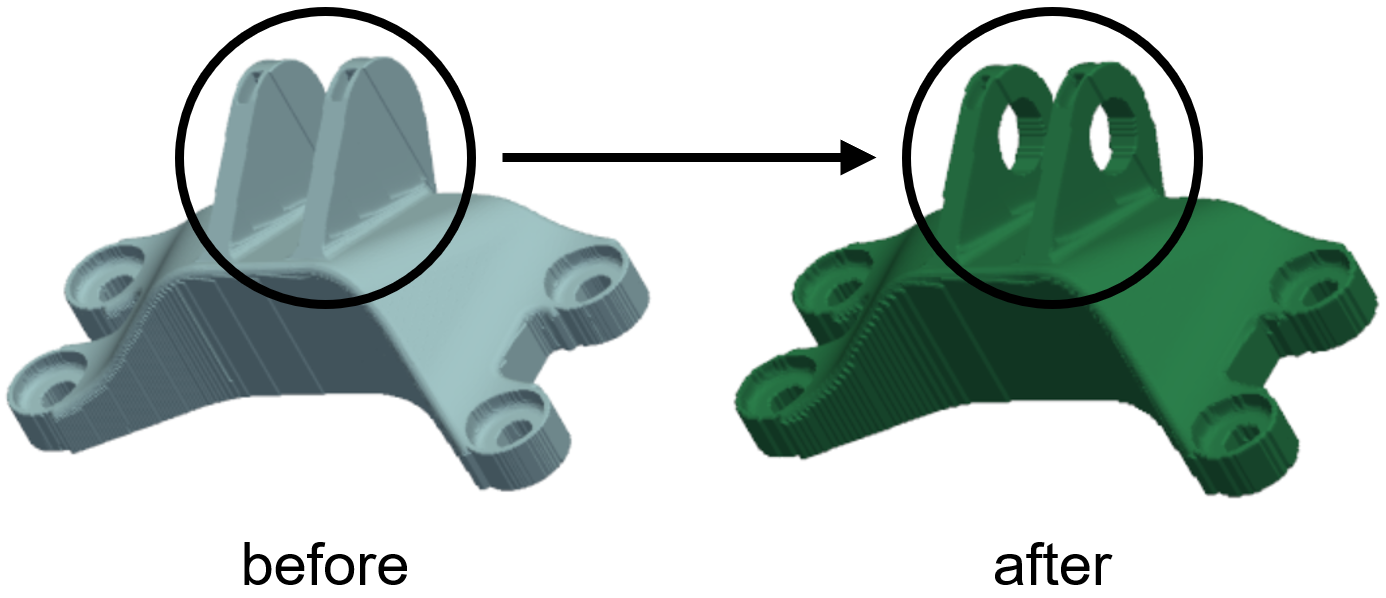}
    \caption{Restoration of transverse mounting holes after depth reconstruction. Boolean subtraction reopens the filled holes (circled).}
    \label{fig:hole-restoration}
\end{figure}

Depth reconstruction fills mounting holes transverse to the selected pull direction.
We restore these specified holes using Boolean mesh operations, as illustrated in Fig.~\ref{fig:hole-restoration}.
In manufacturing, these transverse holes can be formed using side-action core pins that retract along the hole axes before part ejection, or machined after molding.
Their restoration is therefore treated as a separate operation beyond the selected pull-direction guarantee.

\subsection{Shape Generation from Paired Depth Images}\label{sec:generation-method}

Each training shape is converted into a pair of depth images using the procedure above.
A diffusion model learns the distribution of these image pairs and samples new pairs, which are reconstructed into three-dimensional solids by filling the intervals between their boundaries.

The generative-model evaluation used pixel-space diffusion to sample paired depth maps, which were then reconstructed into three-dimensional shapes.
The subsequent FEA-guided optimization study used a separate set of candidates generated by latent-space diffusion.
Both pipelines produce the same paired-depth representation, whose upper and lower boundaries initialize the optimizer.

The pixel-space model operates directly on the $2\times256\times256$ depth tensor, while the latent-space model operates on a $4\times64\times64$ representation produced by a fixed deterministic autoencoder.
Both use a U-Net denoising model with $v$ prediction and a sigmoid noise schedule, followed by 250-step deterministic DDIM sampling \citep{ho2020ddpm,song2020denoising}.
The latent autoencoder was trained from scratch and selected using the validation split.
Its compact spatial representation reduces the workload of diffusion training.
Model settings and computational costs are detailed [See Supplemental Material].

\subsection{FEA-Guided B-Spline Shape Optimization}\label{sec:ffd-method}

A generated depth pair initializes the FEA-guided refinement in Fig.~\ref{fig:framework}(c).
The optimizer changes thickness by moving the upper and lower boundaries along the selected pull direction while holding the in-plane footprint fixed.
Before each FEA evaluation, the boundaries are projected into the design envelope to maintain minimum thickness and prescribed interfaces.
Generation and refinement are sequential: the generative model supplies the initial depth pair, and FEA compliance sensitivities guide subsequent boundary updates.
We refine each boundary by adding a smooth displacement field to the generated depth map,

\begin{equation}\label{eq:ffd}
\mathbf{z}^{+}=\mathbf{z}^{+}_0+B\mathbf{c}^{+},\qquad
\mathbf{z}^{-}=\mathbf{z}^{-}_0+B\mathbf{c}^{-},
\end{equation}

where $\mathbf{z}^{\pm}_0$ are the generated depth maps and $\mathbf{c}^{\pm}$ contain the vertical displacements of an $n_c\times n_c$ grid of control points for each boundary.
The matrix $B\in\mathbb{R}^{HW\times n_c^2}$ maps these control point displacements to the depth map grid using products of clamped-uniform cubic B-spline basis functions in the two in-plane directions.
In this study, we used $n_c=20$, giving 800 scalar control displacements across the two boundaries before fixing the interface controls described below.
Only displacement along $z$ is permitted, and the in-plane sampling grid does not move.

Before each FEA evaluation, the depth boundaries were projected into the design envelope while imposing a prescribed minimum thickness in columns identified as initial material.
The fixed and load interfaces were restored to their prescribed envelope boundaries, and nearby control points were held fixed to preserve these connections.
The same projection settings and 60-iteration optimization budget were used for all candidates.
Thresholds, interface mask construction, and the learning rate schedule are specified [See Supplemental Material].

For analysis, the hard interval in Eq.~\eqref{eq:solid} was replaced by a differentiable density field sampled along $z$,

\begin{equation}\label{eq:density}
\rho_{ijk}=\sigma\!\left(\beta(z_k-z^-_{ij})\right)
\sigma\!\left(\beta(z^+_{ij}-z_k)\right),
\end{equation}

where $\sigma$ is the logistic sigmoid and $\beta$ controls the sharpness of the material boundary.
The density field was transferred to a tetrahedral mesh of the fixed design envelope for linear-elastic FEA.
The mounting interfaces were constrained and a single diagonal load, oriented along $(1,-1,-1)/\sqrt{3}$, was applied at the load interface. This optimization load is distinct from the four load cases used to evaluate the reconstructions (Sec.~\ref{sec:setup}), including their SimJEB \emph{Diagonal} case, which acts along a different direction.
Because the material modulus is normalized and the load is expressed in analysis units, compliance is interpreted as a within-case relative measure.
Density resolution, boundary sharpness, material parameters, mesh and interface definitions, and solver settings are specified [See Supplemental Material].

Compliance sensitivity was propagated through the transpose of the forward density transfer, Eq.~\eqref{eq:density}, and $B^\mathsf{T}$.
The FEA density sensitivity and the linear density-transfer transpose were checked against centered finite differences. This check does not include the hard geometry projections.
Across five random directions at each of three base densities, the adjoint directional derivative matched the centered difference to a relative error below $10^{-3}$ [See Supplemental Material].

Boundary updates combine compliance sensitivities with a volume penalty that discourages material change and a control displacement penalty that limits deformation.
The update is a normalized-gradient heuristic rather than descent on one scalar weighted objective.
Let $g^{+}$ and $g^{-}$ be compliance sensitivities with respect to the projected depth boundaries, $M$ the indicator of columns with current positive thickness, $\mathcal{N}(g)=g/(\lVert g\rVert_2+10^{-12})$, $w_V$ a volume penalty weight (default $3$), $w_c$ a control displacement weight (default $0.3$), and
\begin{align}
V_f&=\frac{1}{HW(z_{\mathrm{hi}}-z_{\mathrm{lo}})}\sum_{i,j}\max(z^+_{ij}-z^-_{ij},0),\\
s&=\tanh\!\left(10(V_f-V_{f,0})\right).
\end{align}
Here $V_{f,0}$ is computed from the envelope-clipped generated depth pair before the per-iteration minimum thickness and interface projections.
The reported initial $V_f$ denotes the step-0 projected state and can therefore differ from $V_{f,0}$.
The control updates are
\begin{align}
q^+&=\mathcal{N}(B^\mathsf{T}g^+)+w_V\,\mathcal{N}(B^\mathsf{T}(sM))+w_c\,c^+,\\
q^-&=\mathcal{N}(B^\mathsf{T}g^-)+w_V\,\mathcal{N}(B^\mathsf{T}(-sM))+w_c\,c^-,\\
c^+&\leftarrow c^+-\eta q^+,\qquad c^-\leftarrow c^--\eta q^-.
\end{align}
The volume and control terms discourage material change and large deformations but do not impose exact constant volume.
Compliance changes are therefore interpreted together with changes in geometric volume, measured from the projected depth boundaries.

The optimizer retains the lowest-compliance projected depth state encountered, including the initial state.
Reported compliance and volume refer to this selected state, which is saved and reanalyzed using the documented checks [See Supplemental Material].
Export meshes undergo separate smoothing and interface restoration and are not reanalyzed by FEA.

\subsection{Evaluation Protocol}\label{sec:setup}

\paragraph{Dataset}\label{ssec:dataset}
We used the SimJEB engine bracket dataset \citep{whalen2021simjeb}.
A fixed cohort of 264 of the 381 public designs shared the study's mounting interface layout.
Reconstruction costs were evaluated on 230 original--reconstruction pairs matched by shape intersection-over-union (IoU), with a separate 150-pair identifier-verified check.
Cohort selection, matching criteria, and the limits of the historical records are detailed [See Supplemental Material].
The diffusion experiments used 211 training, 26 validation, and 27 held-out designs. The held-out set was not used in the generative comparison.
The conversion itself accepts arbitrary surface meshes, but all reported experiments concern this bracket family and the selected coordinate convention.

\paragraph{Representation Evaluation}
The representation analysis used the 230 shape-matched pairs described above.
Its primary mass estimate used material volume on the common 2.5-mm voxel grid; exact mesh volume supplied a discretization cross-check.
Structural response was evaluated using C3D8 voxel models in CalculiX under the four published SimJEB load cases and Ti--6Al--4V material properties.
On an independent set of 1,114 DeepJEB brackets, the same voxel scorer reproduced commercial-solver stiffness rankings with Spearman correlations of 0.88--0.98 across load cases \citep{hong2024deepjeb,dhondt2004calculix}.
This comparison supports ranking use but is not a conforming-mesh validation of every SimJEB reconstruction.
Exact material constants, load vectors, boundary conditions, and interface coupling are provided [See Supplemental Material].

\paragraph{Generative Model Evaluation}
Generation was evaluated across three diffusion seeds, with 256 samples per run, using raw depth validity, reconstructed component counts, and balanced depth distances to the training set and other generated samples.
Depth validity was evaluated before the reconstruction oracle applied its range mapping.
Metric definitions and the exploratory comparison are provided [See Supplemental Material], including a major-volume fragmentation threshold defined after visual inspection.

\paragraph{Shape Optimization Study}
The shape optimization study used 200 candidates generated by latent-space diffusion.
Generated depths were converted to boundary coordinates and clipped to the design envelope before the minimum thickness and interface projections used for FEA.
These prepared states differ from the raw outputs assessed in the generative model evaluation.
A single fixed configuration was used for every candidate, with no per-candidate tuning: identical control lattice size, minimum thickness, iteration count, learning rate schedule, and penalty weights.
Results include all 200 candidates, including those that retained their initial projected states.

OpenAI Codex (OpenAI, desktop application, September 2026) was used for language editing. The authors reviewed and approved all revisions.

\section{Results}\label{sec:results}

\subsection{Reconstruction and Structural Cost}\label{sec:representation-results}
\begin{figure}[tb]
    \centering
    \includegraphics[width=90mm,height=.68\textheight,keepaspectratio]{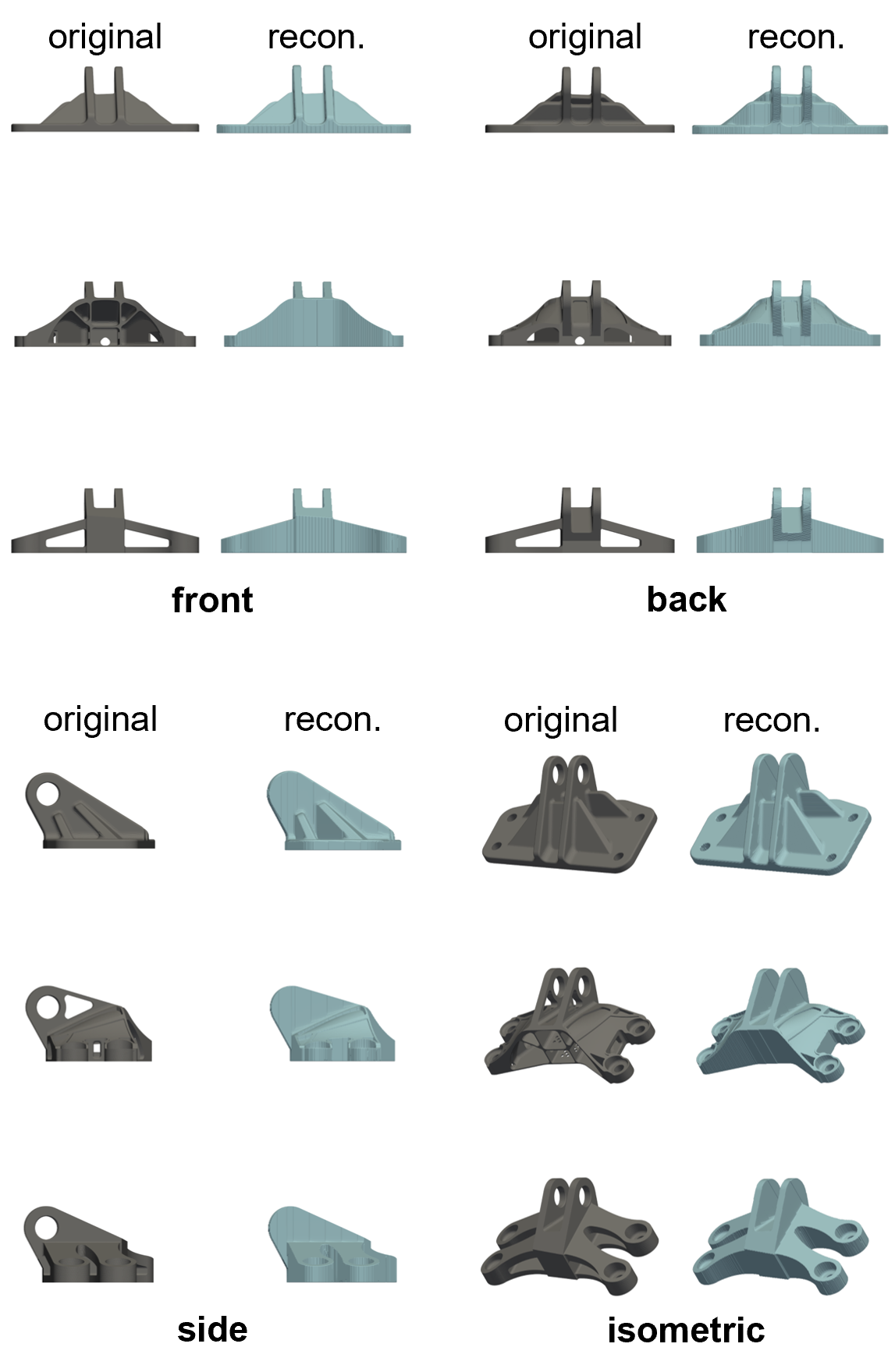}
    \caption{Original brackets and their two-depth reconstructions in front, back, side, and isometric views.}
    \label{fig:results-recon}
\end{figure}

Two-depth reconstruction retains the exposed top and bottom surfaces while filling occluded intervals along the selected pull direction (Fig.~\ref{fig:results-recon}).
The reconstructed brackets therefore preserve their visible outer features but lose internal hollows and recesses that cannot be represented by the two boundaries.

Across the 230 shape-matched pairs, material volume on the common voxel grid increased by a median of $7.8\%$ (interquartile range $4.8$--$11.3\%$).
Under the four SimJEB load cases, median maximum displacement decreased by 13.6--24.4\% relative to the original designs (Table~\ref{tab:structural-cost}).
The interquartile range of displacement changes was entirely negative in every load case.
These results indicate that reconstruction generally increased stiffness while adding material.
Complementary mesh-volume estimates, an identifier-verified subset comparison, and a voxel-resolution check are provided [See Supplemental Material].

\begin{table}[tb]
\centering
\caption{Maximum displacement changes after reconstruction for 230 shape-matched pairs.
Values are median [interquartile range] in percent.\label{tab:structural-cost}}
\small
\begin{tabularx}{\linewidth}{Y c}
\toprule
Load case & Maximum displacement (\%) \\
\midrule
Vertical   & $-16.8\;[-25.5,-11.5]$ \\
Horizontal & $-18.3\;[-26.9,-12.4]$ \\
Diagonal   & $-13.6\;[-22.7,-7.9]$ \\
Torsion    & $-24.4\;[-28.6,-19.8]$ \\
\bottomrule
\end{tabularx}
\end{table}

The resulting mass--stiffness trade-off reflects the filling of occluded regions required by the representation.
For this bracket family, the selected pull-direction restriction was achieved with a median material increase of approximately 8\%.
The accompanying displacement reductions describe the response of the reconstructed designs under the evaluated loads, rather than an improvement in stiffness at equal mass.

\subsection{Generated Designs}\label{sec:generation-results}

\begin{figure}[tb]
    \centering
    \includegraphics[width=.98\linewidth,height=.68\textheight,keepaspectratio]{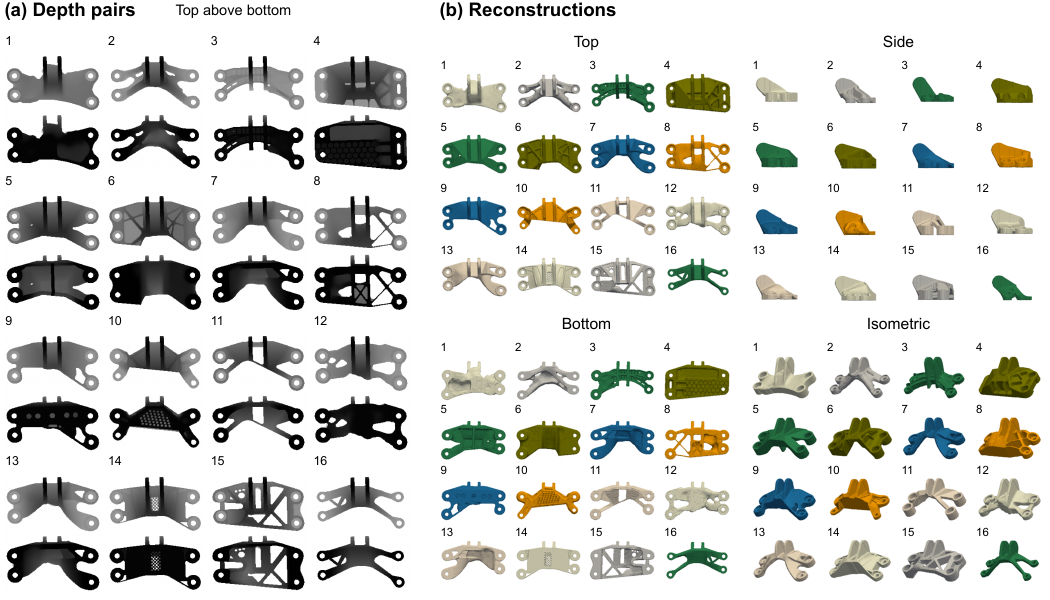}
    \caption{Examples generated by pixel-space diffusion.
    (a) Paired top and bottom depth fields.
    (b) Corresponding three-dimensional reconstructions in multiple views.
    Matching numbers identify the same sample in all views.}
    \label{fig:generated-designs}
\end{figure}

The generated depth pairs reconstruct into three-dimensional bracket candidates with different connecting arms, openings, and rib patterns (Fig.~\ref{fig:generated-designs}).
These examples retain recognizable features of the bracket family while varying the material distribution between the mounting regions.
The paired images describe the shape's outline and its upper and lower surfaces, allowing a two-dimensional generative model to produce three-dimensional design candidates.

The generated depth boundaries also provide the starting variables for structural refinement, linking shape generation to the FEA-guided optimization examined next.
Figure~\ref{fig:generated-designs} shows pixel-space diffusion examples, whereas the optimization study uses candidates from latent-space diffusion.
A separate comparison of their depth validity, connectivity, and sample distances is provided [See Supplemental Material].

\subsection{FEA-Guided Shape Optimization of Generated Designs}\label{sec:ffd-results}

\begin{figure}[t]
    \centering
    \includegraphics[width=95mm,height=.68\textheight,keepaspectratio]{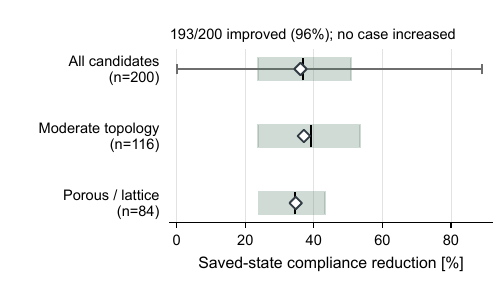}
    \caption{Compliance reduction across 200 generated candidates.
    Bars show the interquartile range, vertical ticks the median, and diamonds the mean.
    The thin line for all candidates shows the full observed range.}
    \label{fig:general}
\end{figure}

FEA-guided refinement reduced compliance in 193 of 200 generated candidates.
Across all 200 candidates, the median reduction was 36.8\% (interquartile range 23.7--50.8\%, range 0--89\%), as summarized in Fig.~\ref{fig:general}.
The remaining seven candidates retained their initial designs.
Figure~\ref{fig:planned-ffd} illustrates the associated shape changes.
Subgroup summaries are reported [See Supplemental Material].

\begin{figure}[t]
    \centering
    \includegraphics[width=.98\linewidth,height=.68\textheight,keepaspectratio]{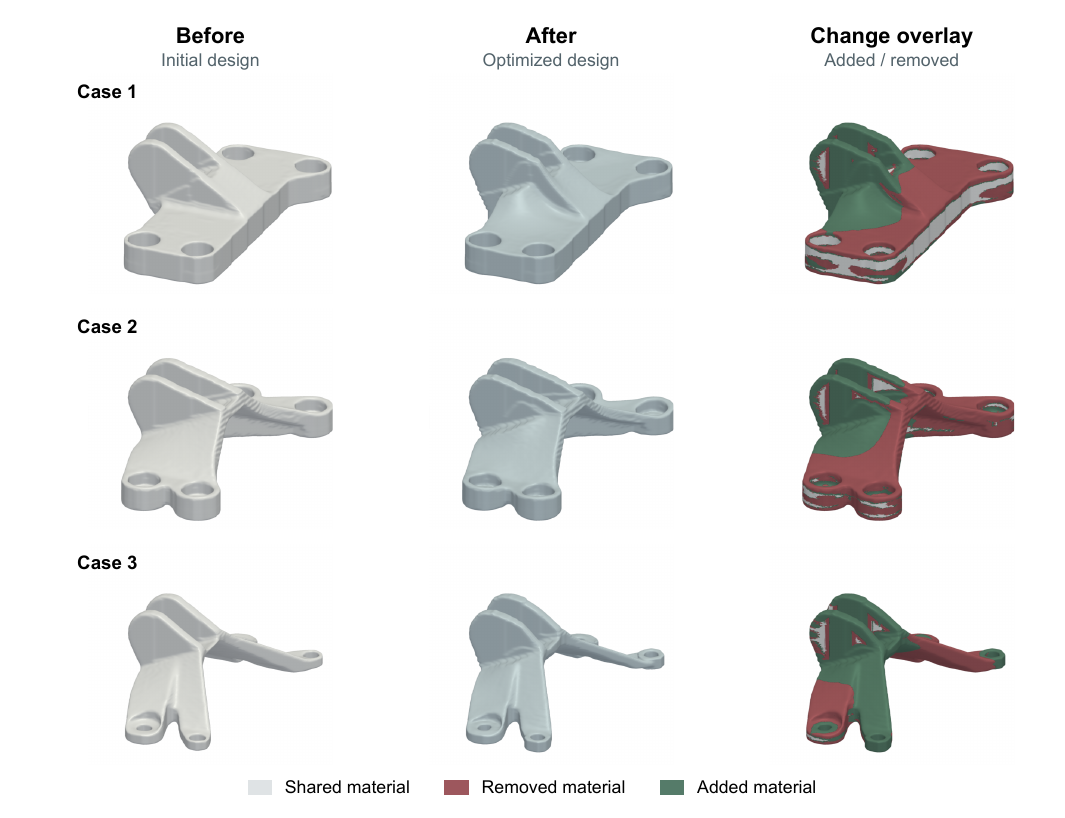}
    \caption{Shape refinement for Cases 1--3. Before shows reconstructions of generated depth fields clipped to the design envelope. After shows the saved optimized designs. Overlays show material added or removed during input preparation and refinement. All surfaces are smoothed for visualization.}
    \label{fig:planned-ffd}
\end{figure}

\begin{figure}[t]
    \centering
    \includegraphics[width=.98\linewidth,height=.68\textheight,keepaspectratio]{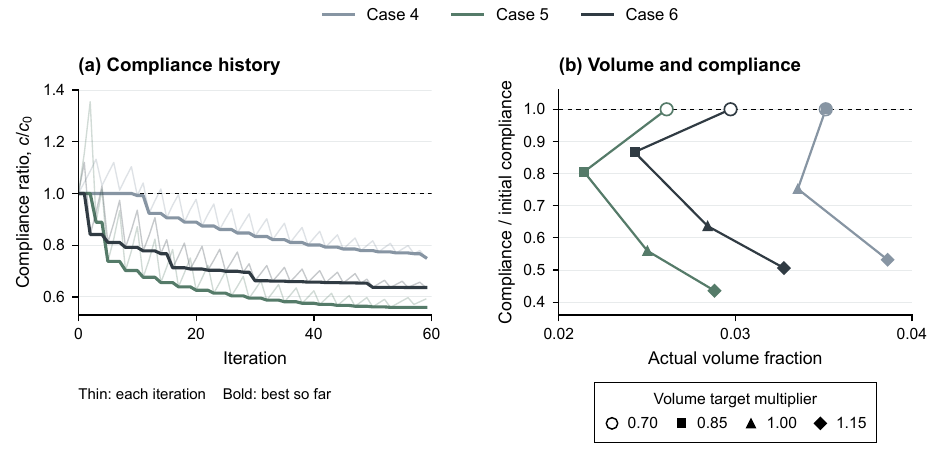}
    \caption{Optimization behavior of three additional candidates (Cases 4--6).
    (a) Compliance histories (thin) and running minima (thick) at the default volume target.
    (b) Volume--compliance outcomes for four volume targets. Two targets retain the same initial design for Case 4.}
    \label{fig:optbehavior}
\end{figure}

Compliance fluctuated during refinement, while the running minimum tracked the lowest compliance reached (Fig.~\ref{fig:optbehavior}a).

To examine the volume--compliance trade-off, we varied the volume target on three additional designs (Cases 4--6) from 0.7 to 1.15 times the reference volume fraction $V_{f,0}$ (Fig.~\ref{fig:optbehavior}b).
Higher volume targets yielded lower or unchanged saved compliance for each design. At factor 0.7, all three runs retained the initial design.
This sweep is an illustrative three-design probe of the trade-off rather than a full frontier estimate.

\begin{figure}[t]
    \centering
    \includegraphics[width=.98\linewidth,height=.68\textheight,keepaspectratio]{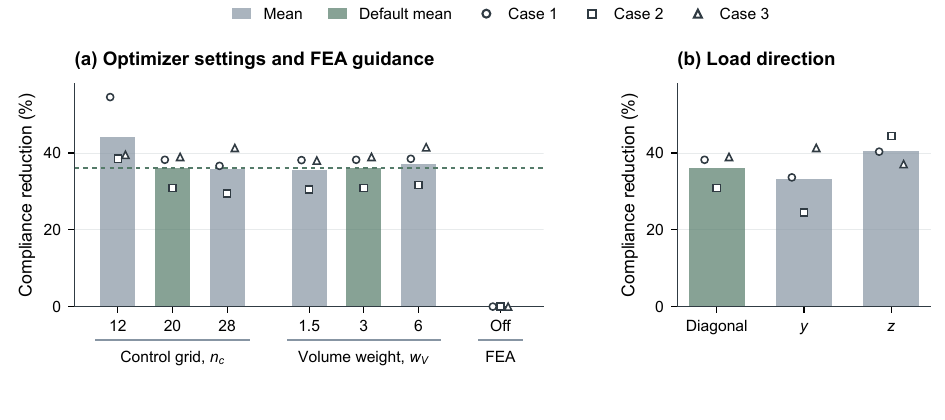}
    \caption{Effects of optimizer settings, FEA guidance, and load direction on compliance reduction for Cases 1--3, the same candidates as in Fig.~\ref{fig:planned-ffd}.
    (a) Effects of control lattice resolution $n_c$, volume penalty weight $w_V$, and removing the FEA gradient (FEA-off).
    (b) Effects of load direction.
    Bars show means, and symbols identify individual candidates. Green bars and the dashed line mark the default configuration.}
    \label{fig:ablation}
\end{figure}

Removing the FEA gradient while retaining the volume and displacement penalties produced no compliance improvement in Cases 1--3 (Fig.~\ref{fig:ablation}a).
With FEA guidance and one setting varied at a time, compliance decreased at each tested control lattice resolution ($n_c\in\{12,20,28\}$: mean $44\%$, $36\%$, $36\%$) and volume penalty weight ($w_V\in\{1.5,3,6\}$: mean $36\%$, $36\%$, $37\%$).
The coarsest lattice gave the largest mean reduction in these three cases.
Separate optimizations under $y$- and $z$-direction loads also reduced compliance, by means of $33\%$ and $41\%$, respectively.

\section{Discussion}\label{sec:discussion}

\subsection{Two-Depth Fields as Shared Design Variables}

Paired depth fields provide a common design space for shape generation and structural refinement. Equation~\eqref{eq:solid} defines a restricted geometric class that can be generated as a two-channel image, reconstructed as a solid, and refined by moving the same upper and lower boundaries in Eq.~\eqref{eq:ffd}. Valid boundary ordering preserves the selected pull-direction restriction throughout these stages, while the quality and variety of candidate shapes depend on the generative model.

Generation and refinement exploit different degrees of freedom within this shared representation. Generation proposes candidates with different footprints and connecting layouts, whereas FEA-guided refinement adjusts the upper and lower boundaries within a fixed footprint. The initial candidate therefore determines the in-plane layout available to subsequent refinement. For the designer, this separates the exploration of alternative layouts from structural improvement through changes in their thickness distribution.

Compliance reductions in 193 of 200 candidates show that a single refinement configuration can improve a range of generated shapes without per-candidate tuning. In the three ablation cases, removing the FEA gradient eliminated the improvement, supporting the contribution of structural sensitivities to refinement. Because the volume penalty does not enforce constant volume, the reported reductions reflect changes in both geometry and material volume and do not establish improved performance at equal mass.

\subsection{Manufacturing Scope}

When existing designs are reconstructed in the paired-depth representation, filling occluded intervals adds material and changes their structural response.
The representation therefore exposes a deliberate mass--stiffness trade-off rather than preserving all internal geometry without cost.
It is best viewed as an early-stage design space for parts governed by a selected pull direction, with later process checks still required.
In use, the designer first selects the pull direction and decides whether occluded voids may be filled.

The selected pull-direction restriction addresses geometric interference during mold separation.
Positive draft remains a separate requirement. The illustrative projection provided in the Supplemental Material was not applied in the reported experiments.
Wall and rib rules, dimensional tolerances, mold filling and solidification, and structural durability require further process-specific assessment.
Features perpendicular to the pull direction, such as a horizontal mounting bore, require secondary machining, a side action, or another manufacturing provision and are not counted as evidence of two-part demolding.

\subsection{Limitations}

The study uses fewer than 300 filtered designs from one bracket family.
The generative evaluation includes three diffusion seeds, with the latent runs sharing one autoencoder.
Together with the observed clustering near training shapes, this limits claims about broad generative diversity.
Whether the representation reduces training data requirements relative to volumetric generation remains to be established through comparisons across training set sizes.

Structural evaluation of the reconstructions uses voxel-based finite element models.
Reported compliance improvements refer to the smoothed-density FEA model used for optimization [See Supplemental Material].
The corresponding sharp-boundary solids were not independently reanalyzed.
The one-factor comparisons cover three candidates and separate load directions rather than simultaneous loading. Comparisons with classical topology optimization or alternative initialization methods remain outside this study.

The in-plane grid limits the smallest resolved feature. Independently of resolution, the single-interval representation cannot preserve internal cavities or features occluded along the selected pull direction.

\section{Conclusion}\label{sec:conclusion}

CAST reformulates three-dimensional shape generation under a selected pull-direction constraint as a two-dimensional image generation problem using paired top and bottom depth fields. Valid depth ordering restricts each in-plane column to a single material interval and excludes undercuts along the selected pull direction.

Two-dimensional generative models proposed bracket geometries, and FEA-guided updates refined the same upper and lower boundaries while retaining the selected pull-direction restriction. Using one fixed configuration, refinement reduced compliance in 193 of 200 generated candidates, with a median reduction of 36.8\% in the optimization FEA model under a volume penalty. Reconstruction of existing SimJEB brackets also exposed a mass--stiffness trade-off: filling occluded intervals added material and was associated with lower displacement across the evaluated load cases.

The geometric guarantee is limited to undercut exclusion along the selected pull direction. Broader manufacturing feasibility requires additional process-specific assessment. CAST thus demonstrates how a manufacturing-oriented representation can simplify generative modeling while retaining a common design space for physics-guided structural refinement.

\section*{Acknowledgment}
This work was supported by grants from the Ministry of Science and ICT (GTL24033-000, N10250154, and No.~2022-0-00986), the Ministry of Trade, Industry and Energy (RS-2025-02317327 and RS-2025-25444634), the Ministry of Oceans and Fisheries (PET0050), and Korea Hydro \& Nuclear Power Co., Ltd. (No.~8-Tech-07).

\section*{Conflict of Interest}
The authors declare no conflicts of interest.

\section*{Data Availability Statement}
The SimJEB dataset is publicly available from its original authors. Data supporting the findings of this study are available from the corresponding author upon request.

\bibliographystyle{unsrtnat}
\bibliography{bibliography}

\end{document}


\maketitle
\suppressfloats[t]
\begin{abstract}
This Supplemental Material provides algorithms, data processing details, model settings, and additional analyses for the paired-depth representation and FEA-guided shape optimization.
Additional analyses compare pixel-space and latent-space diffusion and illustrate a positive draft projection.
\end{abstract}
\keywords{deep generative design, design for manufacturing, die casting, shape representation}

\section{Depth Field Construction and Reconstruction}

\subsection{Orthographic Depth Calculation}
The top and bottom channels store ray distances rather than global $z$ coordinates. The foreground mask is derived from these channels.
Input CAD coordinates in millimeters were mapped by $\widetilde{\mathbf{x}}=s(\mathbf{x}-\mathbf{c})$, with $\mathbf{c}=(17.76,-73.27,32.06)$ mm and $s=9.635\times10^{-3}$ mm$^{-1}$ computed once over the retained 264 designs.
Algorithm~\ref{alg:depthCalc} makes the viewing origins and directions explicit.
A missed ray receives 0.8.
After range mapping, $m_{ij}=1$ only when both depths are below 0.78. A paired foreground column has valid ordering when $D_{\mathrm{top}}+D_{\mathrm{bottom}}\leq0.8$.
Its boundaries are $z^+=0.4-D_{\mathrm{top}}$ and $z^-=-0.4+D_{\mathrm{bottom}}$.

\begin{algorithm}[tb]
\caption{Opposing orthographic depth calculation}\label{alg:depthCalc}
\begin{algorithmic}[1]
\State \textbf{Input:} registered surface mesh $S$; $256\times256$ in-plane grid
\State \textbf{Output:} $D_{\mathrm{top}},D_{\mathrm{bottom}}\in\mathbb{R}^{256\times256}$
\For{each grid index $(i,j)$}
    \State $(x_i,y_j)\gets(2i/256-1,2j/256-1)$
    \State $R_t\gets((x_i,y_j,0.4),(0,0,-1))$
    \State $R_b\gets((x_i,y_j,-0.4),(0,0,1))$
    \State $D_{\mathrm{top}}[i,j]\gets$ nearest nonnegative hit distance of $R_t$ with $S$, or 0.8 if no hit
    \State $D_{\mathrm{bottom}}[i,j]\gets$ nearest nonnegative hit distance of $R_b$ with $S$, or 0.8 if no hit
\EndFor
\State \textbf{return} $D_{\mathrm{top}},D_{\mathrm{bottom}}$
\end{algorithmic}
\end{algorithm}

\subsection{Depth-to-Mesh Reconstruction}
The depth-to-mesh implementation used for the generative experiments is summarized below.
Algorithm~\ref{alg:shapeRecon} mirrors its range mapping and sub-voxel scalar field construction.
Here $\operatorname{divmod}(D,\delta)$ returns integer quotient $q$ and remainder $r$.
The default call uses no Laplacian smoothing.
The scalar values encode the two sub-voxel surface crossings, and the zero level set gives the reconstructed boundary.

\begin{algorithm}[tb]
\caption{Depth-to-mesh reconstruction}\label{alg:shapeRecon}
\begin{algorithmic}[1]
\State \textbf{Input:} $D_t,D_b\in\mathbb{R}^{256\times256}$
\State Replace values $D\ge0.78$ by 0.8 and values $D<0$ by 0
\State $N_z\gets128$; $\delta\gets0.8/N_z$; initialize $S\gets-0.01$ with shape $256\times256\times N_z$
\For{each pixel $(i,j)$}
    \State $(q_b,r_b)\gets\operatorname{divmod}(D_b[i,j],\delta)$; $q_b\gets\min(N_z-1,q_b-1)$
    \State $(q_t,r_t)\gets\operatorname{divmod}(D_t[i,j],\delta)$; $q_t\gets\min(N_z-1,q_t-1)$
    \If{$q_b=N_z-1$ or $q_t=N_z-1$}
        \State $S[i,j,:]\gets0.001$ \Comment{empty-ray column}
        \State \textbf{continue}
    \EndIf
    \For{$\ell=0,\ldots,q_b$}
        \State $S[i,j,\ell]\gets r_b+q_b\delta-\ell\delta$
    \EndFor
    \For{$\ell=0,\ldots,q_t$}
        \State $S[i,j,N_z-q_t-1+\ell]\gets r_t-\delta+\ell\delta$
    \EndFor
\EndFor
\State $M\gets\Call{MarchingCubes}{S,\mathrm{level}=0,\mathrm{spacing}=(1,1,0.8),\mathrm{step}=1}$
\State \textbf{return} $M$
\end{algorithmic}
\end{algorithm}

\subsection{Interface Feature Restoration}
Boolean mesh operations restore specified mounting interfaces after reconstruction.
Features transverse to the selected pull direction require separate tooling or secondary machining.

\section{Dataset Cohorts and Structural Evaluation}
Preprocessing retained a fixed cohort of 264 of the 381 public SimJEB designs with the study's interface layout.
The exact retained-ID list will be released with the public artifacts, but the executable rule that excluded the other 117 meshes was not recovered.
Reconstruction indices did not encode SimJEB identifiers, so candidates were matched to originals on normalized $32^3$ voxel occupancy and then checked after alignment in the common physical frame.
Eligibility required normalized IoU $\geq0.70$ and 2.5-mm world-frame IoU $\geq0.55$.
When multiple candidates claimed one original, the candidate with the highest normalized IoU was retained.
The resulting one-to-one cohort covered 230 distinct originals. The other 34 registered originals had no retained match.
A separate 150-pair subset matched each reconstruction to the same-identifier original and required IoU at least 0.6.
The primary and identifier-verified cohorts gave similar medians: mass change $+7.8\%$ versus $+7.7\%$ and vertical-load maximum-displacement change $-16.8\%$ versus $-17.0\%$.

Both members of each pair were evaluated on the same 2.5-mm voxel grid of eight-node hexahedral elements (C3D8) in CalculiX using Ti--6Al--4V properties ($E=113800$ N/mm$^2$, $\nu=0.342$).
The four bolt regions were joined with \texttt{*RIGID BODY} constraints and all six degrees of freedom of their reference nodes were fixed.
The load interface surface was connected to its reference node using \texttt{*COUPLING} with \texttt{*DISTRIBUTING(1,6)}, and the load was applied at that node.
The same registered interface regions were used for the full cohort.
The four published SimJEB interface load cases are listed in Supp. Table~\ref{tab:loads}.
Refining the voxel pitch from 2.5 to 1.25 mm moved the median changes summarized across the four load cases by approximately 3.6 percentage points.
Exact mesh volumes gave a median material increase of $12.2\%$, compared with $7.8\%$ on the common voxel grid, indicating that discretization affects the estimated magnitude but not the direction of the material change. Structural comparisons used the common-grid volume.

\begin{table}[tb]
\centering
\caption{SimJEB load cases used for original--reconstruction comparison.
Forces are in N and the torsional moment is in N\,mm.\label{tab:loads}}
\small
\begin{tabularx}{\linewidth}{l Y}
\toprule
Case & Interface load \\
\midrule
Vertical & $\mathbf{F}=(0,0,35586)$ \\
Horizontal & $\mathbf{F}=(-37810,0,0)$ \\
Diagonal & $\mathbf{F}=(-28276,0,31404)$ \\
Torsion & $\mathbf{M}=(0,0,564924)$ \\
\bottomrule
\end{tabularx}
\end{table}

\section{Generative Model Configuration}

\subsection{Pixel-Space and Latent-Space Diffusion Configurations}\label{sec:supp-diffusion-config}
The fixed data split contained 211 training, 26 validation, and 27 held-out designs.
The held-out set remained unopened during this comparison.
Both diffusion arms used a U-Net base width of 64 and multipliers $(1,2,4,8)$, $v$ prediction, a sigmoid schedule, 1000 training timesteps, 250-step DDIM sampling with $\eta=0$, Adam with learning rate $10^{-4}$ and betas $(0.9,0.99)$, gradient clipping at 1, no augmentation, 130,000 optimizer steps, and model seeds 17, 29, and 43.
Each run generated 256 samples with the same list of sample-seed identities.
The comparison used raw weights at step 130,000. EMA outputs were used only for diagnostics.

The arms were matched in data order, update exposure, schedule, seeds, and sample count, but they were not a pure encoding ablation.
The pixel-space model operated on $2\times256\times256$ tensors with trajectory clipping enabled.
The latent-space model operated on channel-wise normalized $4\times64\times64$ posterior means with trajectory clipping disabled and included the prior cost of selecting and training the autoencoder.
Input resolution, memory, wall time, and total end-to-end compute therefore differed.
The three latent runs shared one validation-selected autoencoder and therefore represent diffusion-seed variation rather than independent repetitions of autoencoder selection and training.
No matched two-dimensional--three-dimensional comparison was conducted, so this study does not establish a computational advantage over three-dimensional generators.

\subsection{Spatial Autoencoder}
Supp. Table~\ref{tab:ae-config} gives the deterministic spatial autoencoder configuration.
It used neither a discriminator nor a perceptual network, and its posterior mean served as the reconstruction and diffusion input.
Three models were trained from scratch at seeds 17, 29, and 43. The seed-29 model was selected on the 26-design validation split without using the held-out test set.

\begin{table}[tb]
\centering
\caption{Configuration of the deterministic spatial autoencoder used by all three latent diffusion runs.\label{tab:ae-config}}
\small
\begin{tabularx}{\linewidth}{l Y}
\toprule
Item & Setting \\
\midrule
Implementation & diffusers 0.36.0 \texttt{AutoencoderKL}, trained from scratch \\
Input/output & $2\times256\times256$ paired depths; normalization $2x/0.8-1$ \\
Encoder/decoder widths & $(64,128,256)$; two residual layers per block \\
Activation/normalization & SiLU; GroupNorm with 32 groups; mid-block attention \\
Latent & $4\times64\times64$ posterior mean; KL weight 0 \\
Optimization & AdamW, lr $10^{-4}$, betas $(0.9,0.999)$, weight decay 0 \\
Batch/precision & physical batch 8, accumulation 4, effective batch 32; BF16; gradient clip 1 \\
Training/selection & 20,000 optimizer steps; validation every 250; lowest foreground-balanced validation loss \\
Selected checkpoint & seed 29, step 19,750 \\
\bottomrule
\end{tabularx}
\end{table}

\subsection{Training Resource Profile}\label{sec:supp-resource-profile}
Training resources were measured on one NVIDIA A100 80GB GPU with the precision and batch settings in Supp. Table~\ref{tab:resource-profile}.
PyTorch 2.10 \texttt{FlopCounterMode} counted registered forward and backward operations across all microbatches in one optimizer update, excluding optimizer steps and other training overhead.
Two repeated counts agreed exactly. Peak allocated memory was measured over three updates after one warm-up update initialized the optimizer state.
For pixel diffusion, the original two rank-local batches of 16 were combined into one single-process batch of 32. Communication was excluded, so this memory measurement is not the original per-GPU distributed-training peak.

\begin{table}[tb]
\centering
\caption{Single-A100 training resource measurements.
Batch/update lists physical batch size multiplied by gradient accumulation.\label{tab:resource-profile}}
\small
\begin{tabularx}{\linewidth}{l *{6}{>{\centering\arraybackslash}X}}
\toprule
Component & Tensor & Precision & Batch/update & Params [M] & TFLOP/update & Peak [GiB] \\
\midrule
Pixel diffusion & $2\times256\times256$ & FP32 & $32\times2=64$ & 35.716 & 44.976 & 58.466 \\
Latent diffusion & $4\times64\times64$ & FP32 & $32\times2=64$ & 35.723 & 2.800 & 4.769 \\
Autoencoder & $2\times256\times256$ & BF16 & $8\times4=32$ & 13.858 & 26.597 & 8.128 \\
\bottomrule
\end{tabularx}
\end{table}

Including 20,000 autoencoder updates and 130,000 diffusion updates, the latent pipeline accumulated 0.896 EFLOP versus 5.847 EFLOP for pixel diffusion (15.3\%).
These are profiler-attributed operations, combining BF16 autoencoder and FP32 diffusion counts, rather than hardware FLOPs or wall time.
The autoencoder and diffusion model train sequentially, so their peak memory values are reported separately.

\section{FEA-Guided B-Spline Shape Optimization}\label{sec:supp-optimization}
The settings below supplement the boundary update equations in the main article.
Each boundary uses a $20\times20$ clamped-uniform cubic B-spline control lattice.

\subsection{Geometry Projections and Interfaces}
For optimization, generated depths are converted to boundary coordinates with their range clipped to the normalized span. Columns with converted thickness below 0.01 are collapsed to zero thickness, and the boundaries are clipped to the design envelope.
Minimum thickness and interface projections then define the initial FEA state. Raw-depth metrics in Sec.~\ref{sec:supp-geometry-metrics} are evaluated before this preparation and concern the separate generative comparison.
Before every FEA call, the candidate field is clipped to the fixed envelope. Initial material columns with envelope-clipped thickness above 0.006 are retained, a minimum thickness of 0.016 is imposed, and fixed and loaded interface columns are set to the envelope.
The geometry-protection mask is the $z$ projection of \texttt{fixed.stl} and \texttt{load.stl} occupancy voxelized on a $256\times256\times40$ grid.
This voxelizer samples in-plane cell centers at $(2(i+0.5)/256-1,2(j+0.5)/256-1)$, unlike the depth calculation ray grid.
The undilated mask restores the interface columns, while an eight-pixel in-plane dilation selects fixed spline controls.
The fixed configuration uses no separate background mask, 60 updates, initial learning rate 0.02 with linear decay to 0.004, normalized volume update weight 3.0, and control displacement weight 0.3.

\subsection{FEA and Sensitivity Transfer}
The differentiable density uses 40 samples through the normalized $z$ span and sigmoid sharpness 48.
The two-sigmoid product is not multiplied by a foreground mask. When $z^+=z^-=a$, its continuous value at $z=a$ is $1/4$, although the geometric volume of that column is zero.
The influence of this residual density on compliance has not been quantified by a separate sharp-boundary analysis.
The fixed transform maps this density grid to a physical-meter tetrahedral envelope mesh with 0.006-m characteristic size.
The material model uses $E_0=1$, $\nu=0.3$, SIMP exponent 3, and $E_{\min}=10^{-6}$.
With the default vertex-distance setting, mesh nodes whose nearest \texttt{fixed.stl} vertex lies within 0.006 m receive Dirichlet constraints.
A P1 nodal indicator over mesh nodes within 0.006 m of the nearest \texttt{load.stl} vertex is integrated over the domain and scaled to a total load magnitude of 42,300 analysis units in direction $(1,-1,-1)/\sqrt{3}$.
These solver regions are distinct from the voxel-projected geometry masks.
The conjugate gradient solve uses algebraic multigrid, relative tolerance $10^{-9}$, and at most 3000 iterations.

The forward transfer lets each FEA node select its nearest external density sample and assigns each tetrahedron the mean of its four nodal densities.
Its backward pass is the exact transpose of that linear transfer.
The FEA density sensitivity and linear density transfer were checked with centered finite differences along five random directions at each of three base densities (0.3, 0.5, and 0.7).
The maximum relative error was below $10^{-4}$ at every base density.
Hard envelope, interface, and minimum thickness projections are evaluated before FEA but are not included in that derivative check.

\subsection{Saved Designs and Export}
The projected arrays with the lowest finite compliance are retained, including the initial state.
The selected arrays are saved, reloaded with elementwise exact identity, and reanalyzed. A run fails if compliance differs by relative tolerance more than $10^{-6}$.
Compliance and $V_f$ therefore describe this projected depth state.
Exported meshes are separate visualization and geometry check artifacts: they use 200 hard occupancy layers, Gaussian signed distance smoothing with $\sigma=1.4$, removal of components with at most 200 faces, ten Taubin-smoothing iterations, and analytic restoration of the axial bolt holes and the secondary horizontal bore.
Watertightness and component count describe that export. Its structural response is not reanalyzed.

\begin{table}[tb]
\centering
\caption{Reported optimization results for 200 generated candidates under one fixed configuration.
Reduction refers to compliance of the selected projected depth state.
``Single-connected'' counts projected depth footprints with one component.
\label{tab:supp-ffd}}
\small
\begin{tabularx}{\linewidth}{Y *{6}{>{\centering\arraybackslash}X}}
\toprule
Group & $n$ & Improved & Mean [\%] & Median [\%] & IQR [\%] & Single-connected \\
\midrule
All candidates & 200 & 193 & 36.1 & 36.8 & 23.7--50.8 & 195 \\
\;\;\;moderate topology & 116 & 112 & 37.1 & 39.1 & 23.7--53.5 & 116 \\
\;\;\;porous / lattice & 84 & 81 & 34.7 & 34.5 & 23.9--43.3 & 79 \\
\bottomrule
\end{tabularx}
\end{table}

Supp. Table~\ref{tab:supp-ffd} provides the optimization results by candidate group.

\section{Additional Analyses and Examples}

\subsection{Geometry Metrics}\label{sec:supp-geometry-metrics}
Raw depths were scored before external repair.
The out-of-range fraction is the fraction of all values below 0 or above 0.8.
A channel pixel is foreground when its depth is below 0.78.
Crossing is evaluated only where both channels are foreground and records the fraction satisfying $D_{\mathrm{top}}+D_{\mathrm{bottom}}>0.8$.
The fixed oracle then applies its own range mapping before reconstruction.

For two depth pairs $A$ and $B$, the balanced absolute distance first averages absolute error separately over foreground and background pixels for each channel and each direction, then symmetrizes the $A$-masked and $B$-masked values.
A comparison is null if either pair lacks foreground or background in any channel.
Nearest-training distance is the minimum of this symmetric distance over the 211 training designs; smaller values mean greater proximity to the training set.
The post-hoc within-run nearest-neighbor distance uses the same metric and excludes the sample itself; smaller values indicate denser local clusters.

Raw oracle meshes were split after degenerate faces were removed.
Component count includes every resulting component.
The post-hoc major-volume flag is one when the component with the most faces contains less than 99\% of total absolute component volume.
Largest-component renderings are diagnostic extracts and do not replace the raw generator output.

\subsection{Pixel-Space and Latent-Space Comparison}\label{sec:supp-pixel-latent-results}

\begin{figure}[t]
    \centering
    \includegraphics[width=.98\linewidth,height=.68\textheight,keepaspectratio]{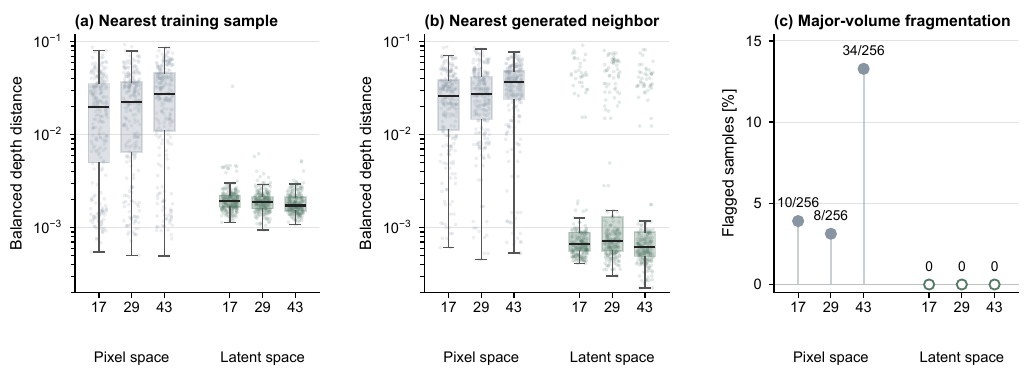}
    \caption{Pixel-space and latent-space diffusion across three seeds, with 256 samples per run.
    Balanced depth distance to (a) the nearest training design and (b) the nearest other generated sample.
    Boxes show medians and interquartile ranges, whiskers extend to 1.5 times the interquartile range, and points show samples.
    (c) Major-volume fragmentation, with flagged counts. Metrics are defined in Sec.~\ref{sec:supp-geometry-metrics}.}
    \label{fig:diffusion-comparison}
\end{figure}

Supp. Fig.~\ref{fig:diffusion-comparison} and Supp. Table~\ref{tab:pixel-latent} report the three-seed comparison.
Latent outputs triggered no major-volume fragmentation flags, but had more small components, more depth-range and crossing violations, and smaller distances to training designs and other generated samples.
No run contained exact duplicate depth arrays.
The fragmentation threshold was defined after inspection, and the difference was concentrated in one pixel-space seed. These results describe the evaluated configurations and do not establish general superiority of either pipeline.

\begin{table}[t]
\caption{Pixel-space and latent-space diffusion results, with 256 samples per run.
Out-of-range fractions, crossing fractions, and nearest-training distances are run means.
Metrics are defined in Sec.~\ref{sec:supp-geometry-metrics}.\label{tab:pixel-latent}}
\centering
\small
\begin{tabularx}{\linewidth}{l *{6}{>{\centering\arraybackslash}X}}
\toprule
Method & Seed & Major-volume flags & Median components & Out of range [\%] & Crossing [\%] & Nearest-train distance \\
\midrule
Pixel space & 17 & 10/256 & 5  & 0.00 & 0.146 & 0.02248 \\
Pixel space & 29 & 8/256  & 9  & 0.00 & 0.117 & 0.02498 \\
Pixel space & 43 & 34/256 & 4  & 0.00 & 0.171 & 0.03046 \\
Latent space & 17 & 0/256 & 15 & 2.93 & 0.450 & 0.00216 \\
Latent space & 29 & 0/256 & 11 & 3.17 & 0.326 & 0.00197 \\
Latent space & 43 & 0/256 & 10 & 3.74 & 0.313 & 0.00187 \\
\bottomrule
\end{tabularx}
\end{table}

Latent diffusion supplied the optimization candidates because of its lower measured training cost (Sec.~\ref{sec:supp-resource-profile}), despite these geometry trade-offs.

\subsection{Illustrative Positive Draft Projection}
\begin{figure}[tb]
\begin{minipage}[t]{0.48\textwidth}
\vspace{0pt}
\begin{algorithm}[H]
\caption{Illustrative discrete positive draft projection for one depth surface}\label{alg:draftProjection}
\begin{algorithmic}[1]
\State \textbf{Input:} depth surface $D_1$, opposing surface $D_2$, draft angle $\theta$, resolution $R$
\State $\Delta x\gets 2/R$; $\delta\gets\cot(\theta)\Delta x$; $k\gets0$
\Repeat
    \State $k\gets k+1$; compute finite difference gradients of $D_1$
    \State $S\gets\{p:\lVert\nabla D_1(p)\rVert>\cot(\theta)\}$
    \State $(D'_1,D'_2,n_{\mathrm{update}})\gets(D_1,D_2,0)$
    \For{each $p\in S$}
        \State $q\gets$ valid cardinal neighbor in the direction of steepest ascent
        \If{$q$ is foreground and the available thickness at $q$ is at least $\delta$}
            \State $D'_1(p)\gets D_1(q)-\delta$; $D'_2(p)\gets D_2(q)$
            \State $n_{\mathrm{update}}\gets n_{\mathrm{update}}+1$
        \EndIf
    \EndFor
    \State $(D_1,D_2)\gets(D'_1,D'_2)$
\Until{$n_{\mathrm{update}}=0$ or $k=1000$}
\State \textbf{return} $D_1,D_2$
\end{algorithmic}
\end{algorithm}
\end{minipage}\hfill
\begin{minipage}[t]{0.48\textwidth}
\vspace{0pt}
    \centering
    \includegraphics[width=.62\linewidth,height=.68\textheight,keepaspectratio]{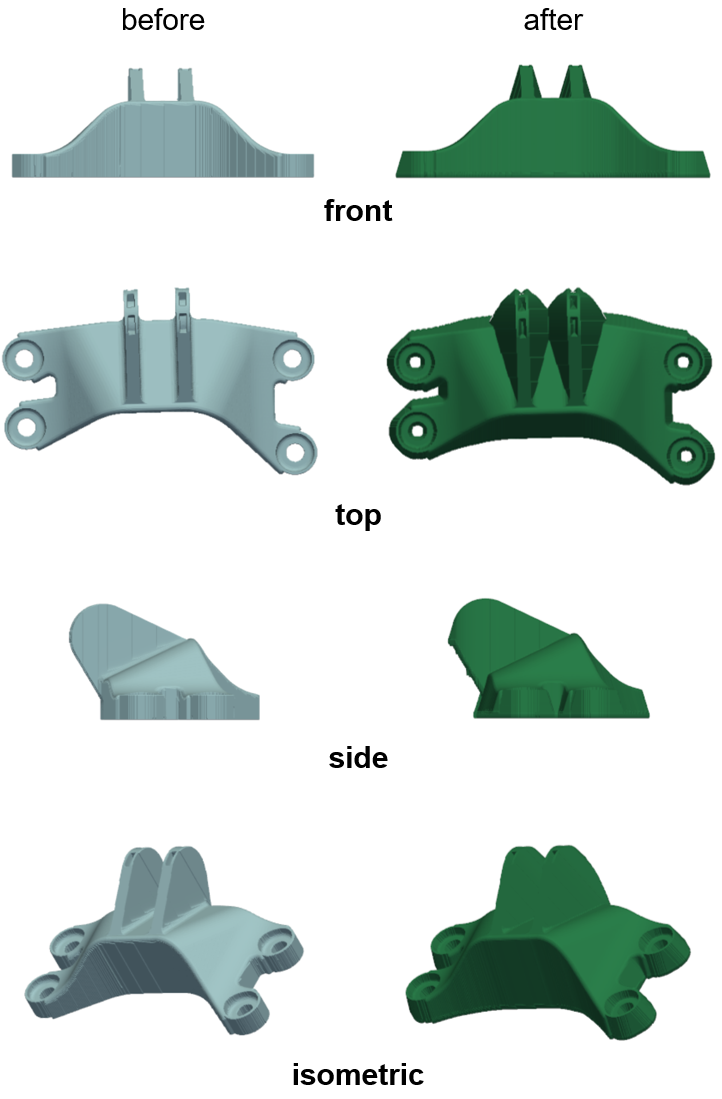}
    \caption{Before and after an illustrative draft projection with an exaggerated angle. This operation was not applied to the reported experiments.}
    \label{fig:supp-draft-angle}
\end{minipage}
\end{figure}

For a depth field sampled with in-plane pitch $\Delta x$, a draft angle $\theta$ measured from the pull axis corresponds to the local finite difference bound
\begin{equation}
\frac{|D_p-D_q|}{\Delta x}\leq\cot(\theta)
\end{equation}
for neighboring pixels $p$ and $q$.
Algorithm~\ref{alg:draftProjection} adjusts neighboring depth values toward this bound while checking the opposing field for available thickness.
Supp. Fig.~\ref{fig:supp-draft-angle} shows the geometric effect with an exaggerated angle.
This illustrative operation was not applied to the reported experiments.

